\documentclass[10pt,twocolumn,letterpaper]{article}

\usepackage{iccv}
\usepackage{times}
\usepackage{epsfig}
\usepackage{graphicx}
\usepackage{amsmath}
\usepackage{amssymb}
\usepackage{booktabs}
\usepackage{comment}
\usepackage{booktabs}
\usepackage{multirow}
\usepackage{float}
\usepackage{authblk}

\usepackage{url}
\usepackage{epsfig}
\usepackage[np, autolanguage]{numprint}
\usepackage{enumitem}
\usepackage[utf8]{inputenc}
\usepackage{soul}
\usepackage{booktabs}
\usepackage[dvipsnames]{xcolor}
\definecolor{BN1}{RGB}{255,176,0}
\definecolor{BN9}{RGB}{254,97,0}
\definecolor{BN17}{RGB}{220,38,128}
\definecolor{BN33}{RGB}{120, 93, 240}

\usepackage{xcolor}

\graphicspath{{figures/}}

\newcommand{\fig}[1]{Fig.~\ref{fig:#1}}
\newcommand{\tab}[1]{Table~\ref{tab:#1}}

\usepackage{graphicx}
\usepackage{amsmath}
\usepackage{amssymb}
\usepackage{booktabs}
\usepackage{comment}

\usepackage{mathtools}

\usepackage[pagebackref=true,breaklinks=true,letterpaper=true,colorlinks,bookmarks=false]{hyperref}

\iccvfinalcopy % *** Uncomment this line for the final submission

\def\iccvPaperID{****} % *** Enter the ICCV Paper ID here

\ificcvfinal\fi

\makeatletter
\renewcommand\AB@affilsepx{, \protect\Affilfont}
\makeatother

\begin{document}

%%%%%%%%% TITLE
\title{Video BagNet: short temporal receptive fields\\increase robustness 
in long-term action recognition}
\author[*,1,2]{Ombretta Strafforello}
\author[*,1]{Xin Liu}
\author[2]{Klamer Schutte}
\author[1]{Jan van Gemert}
\affil[*]{Authors with equal contribution}
\affil[1]{Delft University of Technology} 
\makeatletter
\renewcommand\AB@affilsepx{\\ \protect\Affilfont}
\makeatother

\affil[2]{TNO}

\affil[ ]{\textit{\{o.strafforello, x.liu-11, j.c.vangemert\}@tudelft.nl, \{klamer.schutte\}@tno.nl}}

\renewcommand\Authands{ and }

\maketitle
\ificcvfinal\thispagestyle{empty}\fi

%%%%%%%%% ABSTRACT

%\squeezeup

\begin{abstract}

Previous work on long-term video action recognition relies on deep 3D-convolutional models that have a large temporal receptive field (RF). 
We argue that these models are not always the best choice for temporal modeling in videos. 
A large temporal receptive field allows the model to encode the exact sub-action order of a video, which causes a performance decrease when testing videos have a different sub-action order.
% In this work we introduce the Video BagNet, a variant of the 3D-ResNet50 model with the temporal receptive field size limited to 9 or 33.
In this work, we investigate whether we can improve the model robustness to the sub-action order by shrinking the temporal receptive field of action recognition models. %We do \klamer{For this we employ (instead of We do this by employing)} this by employing 
For this, we design Video BagNet, %\jvg{'Video-BagNet' instead of 'Video BagNet' ?} 
a variant of the 3D ResNet-50 model with the temporal receptive field size limited to 1, 9, 17 or 33 frames. 
We analyze Video BagNet on synthetic and real-world video datasets and experimentally compare models with varying temporal receptive fields. We find that short receptive fields are robust to %classify \rb{classifying?} 
sub-action order changes, while larger temporal receptive fields are sensitive to the sub-action order.
\end{abstract}

%\squeezeup 

%%%%%%%%% BODY TEXT
\begin{figure*}[!ht]
	\centering
	\begin{tabular}{c@{}c}
	\includegraphics[width= 0.45\linewidth]
	{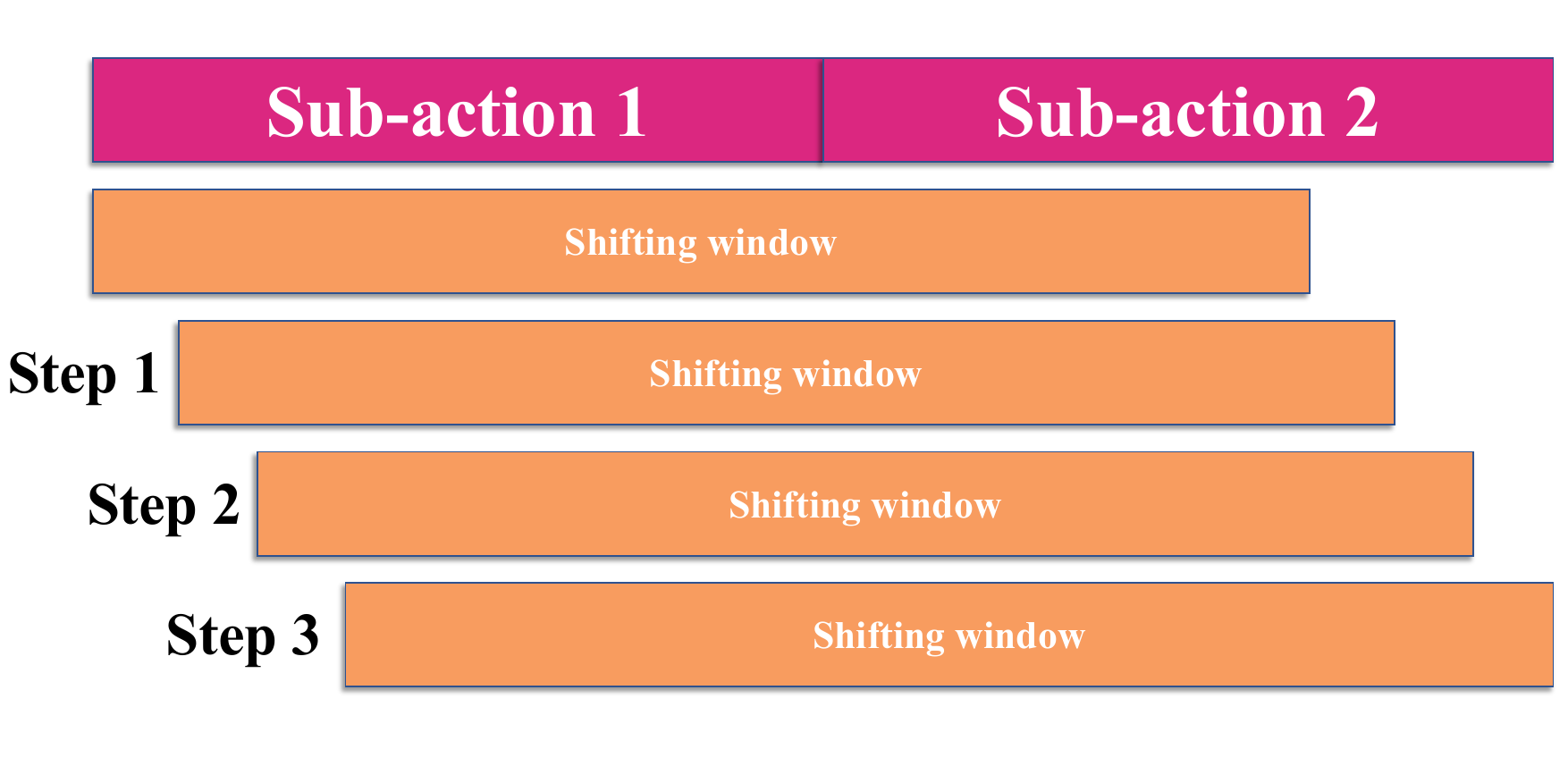} & 
	\includegraphics[width= 0.45\linewidth]
	{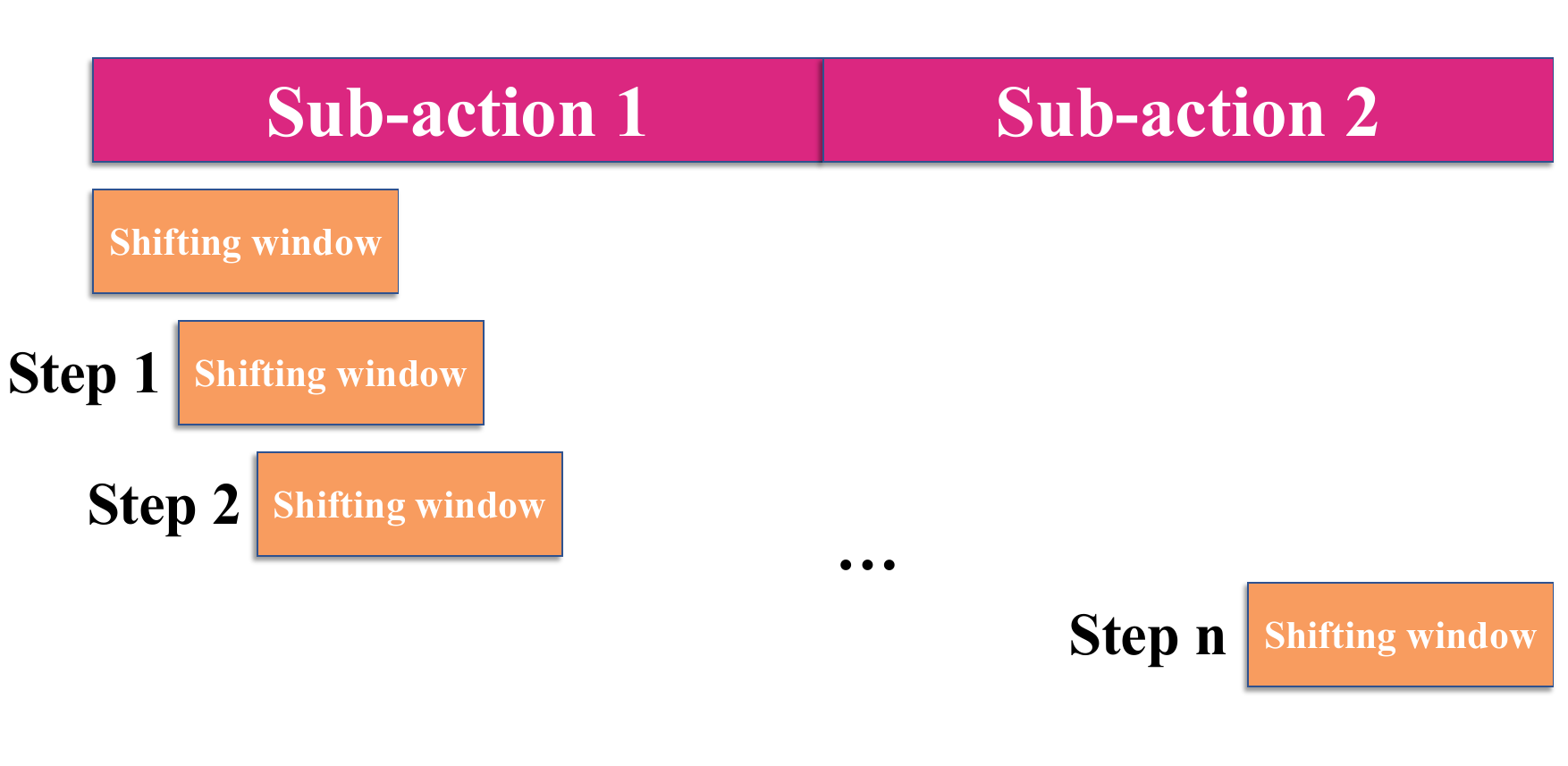} \\
	(a) Model with large temporal RF & (b) Model with small temporal RF\\
	\end{tabular}
	\caption{Large (a) versus small (b) temporal RF compared to the sub-action duration. The temporal RF size in the last convolutional layer is represented by the size of the convolutional shifting windows.
	Models with large temporal RF see sub-actions in ordered co-occurrences, while models with small temporal RF are more likely to see single sub-action occurrences. Because of this, models with small temporal RFs encode sub-action occurrences but not strict sub-action orders.%\klamer{this figure lacks a notion of a final pooling layer which summarizes over a whole video, shouldn't that be added}\om{todo:add global avg pooling}
 }
	\label{fig:tRF_comparison}
\end{figure*}

\section{Introduction}
\label{sec:introduction}

Long-term action videos naturally have %sub-action permutations where videos of the same action class contain 
different sub-action combinations and orders.
For instance, the action of 'making coffee' may contain either order of 'add sugar, add milk', or 'add milk, add sugar', or people can drink their coffee black. With such diversity in sub-action orders it is nearly impossible to sample representative data containing all possible permutations for training a long-term action recognition classifier. Thus, the training set in current long-term classification datasets
like MultiTHUMOS~\cite{MultiTHUMOSDataset} and Charades~\cite{Charades}  may contain different sub-action orders than the test set. The specific sub-action order and duration is exploited by current video action recognition models due to their large temporal receptive field size.
Consequently, if the models encode the specific sub-action order at training time, it might cause misclassification of a video action when the sub-action order differs at test time. 

In this paper, we focus on encoding sub-action \textit{order}. We refer to the \textit{temporal receptive field} (RF) as the number of input frames within a shifting kernel that a network can make use of in its last convolutional layer. 
Usually, the last convolutional layer is followed by global temporal pooling, which collapses the temporal dimension into one unit, and a final fully connected layer. These operations do not affect the temporal RF size and the sensitivity to order, as they cannot model temporal dependencies. For this reason, we do not consider the final pooling and classification layers in our calculation of the temporal RF size.
Networks with temporal RF size larger than the sub-action duration (as shown in \fig{tRF_comparison} (a)) might overfit on the exact sub-action order seen at training time. 
In cases where the available training samples are not sufficiently representative of all possible sub-action orders, misclassifications occur at test time.

We introduce Video BagNet, a model with a small temporal RF size that is less sensitive to the exact sub-action order.
Our model is inspired by BagNet~\cite{BagNet}, which reduces the spatial receptive field size for easier network interpretation.
We %utilize \rb{use, see also Jan's guidelines :)} \jvg{Nice! :D} 
use Video BagNet to investigate the role of the temporal RF in encoding the sub-action order.
Our proposed Video BagNet is modified from 3D ResNet-50 \cite{hara2018can}. %\rb{cite?}.
%We reduce the temporal receptive field size by making kernels smaller \klamer{shrinking the kernels} on the temporal dimension , and using fewer \klamer{less} down-sampling.
We reduce the temporal RF size by shrinking the kernels in the temporal dimension and using less down-sampling.
As shown in \fig{tRF_comparison} (b), %our Video BagNet with small temporal receptive field size are \klamer{is (or use a plural form for the Video BagNet)}
our Video BagNet with small temporal RF sizes is less sensitive to the exact sub-action order by seeing 
%more \klamer{delete more} occurrences of single sub-actions \klamer{rather} 
occurrences of single sub-actions rather than the combinations of ordered sub-actions.
This results in better sub-action detection performance than 3D ResNet-50 on our synthetic \textit{Directional Moving MNIST} dataset and MultiTHUMOS.
We also provide a measurement of model sensitivity to the sub-action order. % in this paper.
% \\\\
% \om{Mention that RF also encodes length but we do order.}
%
Our code will be made publicly available\footnote{\url{https://github.com/ombretta/videobagnet}}.

\section{Related Work}
\label{sec:related_work}

\subsection{Temporal extent of recent models for action recognition}

Recent action recognition architectures can model long temporal extents \cite{hussein2019timeception, hussein2020timegate, liu2021no, qiu2019learning, wang2018non, DBLP:journals/corr/abs-1812-05038, zhou2018temporal}. This is achieved through two main approaches. The first one is by extending the temporal receptive field of convolutional models, either by stacking strided convolutional layers, thus making the model deeper \cite{carreira2017quo, tran2015learning}, or by harnessing auxiliary temporal modules \cite{hussein2019timeception, hussein2019videograph, wang2018non}. %\jvg{Another example of (?)} 
The second approach is by means of transformer architectures, whose design entails a temporal receptive field which spans over the whole input duration \cite{DBLP:journals/corr/abs-2102-05095, patrick2021keeping, DBLP:journals/corr/abs-2103-13915}. Large temporal extents make it possible to learn dependencies in videos over time. This allows for modeling the order of the sub-actions that are seen at training time, which is considered useful to capture the inner structure of complex, long-term activities \cite{hussein2020pic}.

However, models with large temporal RF have a drawback: they 
% have a very large number of parameters that make them
are prone to overfitting on the order when the available training data is limited \cite{hara2017learning}. 
%\xin{I do not think we should put 'parameters' here. It is not related to encode sub-action orders and makes confusion.}
This is the case for most of the current long-term action recognition datasets, which only consist of a few hundred or thousand videos \cite{BreakfastDataset, sigurdsson2016hollywood, MultiTHUMOSDataset}.
In addition, recent work showed that some of the current long-term action recognition datasets can be solved without using long-term information \cite{strafforello2023longterm}.
In this work, we investigate whether modeling large temporal extents is always beneficial to solve long-term action recognition. In particular, we investigate whether models with large temporal RF overfit on the order of the sub-actions seen at training time, causing misclassifications at test time.

\subsection{Order invariant networks}
%\jvg{Remove: Sensitivity to sub-action order in long-term action recognition models has been researched before.} 
In \cite{hussein2019videograph}, it is empirically shown that the classification performance of order-aware methods drops significantly when new sub-action orders are presented at test time. On the other hand, order invariant methods, like ActionVLAD \cite{girdhar2017actionvlad}, are robust to sub-actions permutations.
Hussein \etal \cite{hussein2020pic} propose a permutation invariant convolutional module, PIC, to model temporal dynamics in long-range activities. The PIC module performs self-attention across pre-extracted visual features and can be stacked on top of convolutional backbones. PIC is robust to sub-action permutation compared to ordered-aware convolutional baselines \cite{hussein2019timeception}, while maintaining a large temporal RF. 

Our approach deviates from ActionVLAD and PIC. While ActionVLAD is completely order unaware, we maintain order information within short receptive fields. This allows modeling fine-grained motions, which is proven beneficial for action recognition \cite{huang2018makes, sigurdsson2017actions}. Differently than PIC, we investigate sensitivity to sub-action order by looking at the temporal RF size of spatio-temporal convolutional networks, commonly used as backbones in long-term action recognition models \cite{hussein2019timeception, hussein2019videograph, wang2018non}. Our method only requires simple modification to the spatio-temporal convolutional networks.

\subsection{Reducing the receptive field size: BagNet}

Our idea of reducing the temporal receptive field size is inspired by Brendel \etal \cite{BagNet}, who investigated how bag-of-local-features can be used for image classification. Bag-of-local-features can be obtained by restricting the spatial receptive field of the image classifier to a small number of pixels. %In the proposed \klamer{In Brendel's} 
In Brendel \etal's model, the \textit{BagNet},
%which is an approximation of ResNet-50 \cite{DBLP:journals/corr/HeZRS15}, 
this is achieved by replacing a set of $3\times3$ convolutions with $1\times1$ convolutions and removing the first downsampling layer. The property of this architecture is that the image feature representation is given by a collection of local features, corresponding to small image patches, that do not take into account the global spatial structure. Surprisingly, ignoring global structures does not hurt substantially the classification accuracy of BagNet.
%The advantage of BagNet is in its interpretability: it is possible to directly trace which patch determined the classification outcome. \klamer{we do nothing with this advantage? So perhaps rephrase as 'An additional advantage ...'}
%An additional advantage of BagNet is in its interpretability: it is possible to directly trace which patch determined the classification outcome. 
%\jvg{Conclusion of paragraph missing?}
Using bag-of-local-features has been taken on for other visual classification tasks. Some examples are exploring local features for face anti-spoofing \cite{Shen_2019_CVPR_Workshops}, and predicting the histogram of visual words of a discretized image as part of a self-supervision task \cite{gidaris2020learning}. To the best of our knowledge, our method is the first work that relies on % bag-of-local-features \klamer{bag-of-temporal-features} 
bag-of-temporal-features models to learn video representations.

\begin{table*}[!ht]
\centering
\begin{tabular}{lccc}
         %\thickhline 
         & \textit{3D ResNet-50 (\textit{RN})}     & \textit{Video BagNet-\textcolor{BN1}{1}/\textcolor{BN9}{9}/\textcolor{BN17}{17}/\textcolor{BN33}{33} (\textit{BN})}     &\\ %\hline
         
{\begin{tabular}[c]{@{}l@{}} 
\# parameters\\
for 3 classes
\end{tabular}}   & 46.2 M & \textcolor{BN1}{45.9}/\textcolor{BN9}{46.7}/\textcolor{BN17}{45.6}/\textcolor{BN33}{46.5} M &  Output sizes $T\times S^2$\\ \hline

conv1   & $7\times7^2, 64$, stride (1, 2, 2) & $\textcolor{BN1}{1}/\textcolor{BN9}{3}/\textcolor{BN17}{3}/\textcolor{BN33}{3}\times7^2, 64\times k$, stride (1, 2, 2) & 
{\begin{tabular}[c]{@{}l@{}} 
\textit{RN}: $64\times32^2$\\
\textit{BN}: $64\times32^2$
\end{tabular}} \\ \hline

downsampling    & Max pool (3, 3, 3), stride 2 & 
Max pool (1, 3, 3), stride (1, 2, 2) & 
{\begin{tabular}[c]{@{}l@{}} 
\textit{RN}: $32\times16^2$\\
\textit{BN}: $62\times16^2$
\end{tabular}}\\ \hline

conv2\_x & 
\begin{tabular}[c]{@{}l@{}} $ \begin{bmatrix}
1\times1^2, 64\\
3\times3^2, 64\\
1\times1^2, 64
\end{bmatrix} $, \\
$ \begin{bmatrix}
1\times1^2, 256\\
3\times3^2, 64\\
1\times1^2, 64
\end{bmatrix} \times 2 $
\end{tabular} & 
\begin{tabular}[c]{@{}l@{}} $ \begin{bmatrix}
1\times1^2, 64\times k  \\
\textcolor{BN1}{1}/\textcolor{BN9}{3}/\textcolor{BN17}{3}/\textcolor{BN33}{3}\times3^2, 64\times k \\
1\times1^2, 64\times k 
\end{bmatrix} $, \\
$ \begin{bmatrix}
1\times1^2, 256\times k \\
\textcolor{BN1}{1}/\textcolor{BN9}{1}/\textcolor{BN17}{1}/\textcolor{BN33}{1}\times3^2, 64\times k \\
1\times1^2, 64\times k 
\end{bmatrix} \times 2 $
\end{tabular} & 
{\begin{tabular}[c]{@{}l@{}} 
\textit{RN}: $32\times16^2$\\
\textit{BN}: $60\times16^2$
\end{tabular}}\\ \hline

conv3\_x &
\begin{tabular}[c]{@{}l@{}} $ \begin{bmatrix}
1\times1^2, 256\\
3\times3^2, 128\\
1\times1^2, 128
\end{bmatrix} $, \\
$ \begin{bmatrix}
1\times1^2, 512\\
3\times3^2, 128\\
1\times1^2, 128
\end{bmatrix} \times 3 $
\end{tabular} &
\begin{tabular}[c]{@{}l@{}} $ \begin{bmatrix}
1\times1^2, 256\times k \\
\textcolor{BN1}{1}/\textcolor{BN9}{3}/\textcolor{BN17}{3}/\textcolor{BN33}{3}\times3^2, 128\times k \\
1\times1^2, 128\times k 
\end{bmatrix} $,  \\
$ \begin{bmatrix}
1\times1^2, 512\times k \\
\textcolor{BN1}{1}/\textcolor{BN9}{1}/\textcolor{BN17}{1}/\textcolor{BN33}{1}\times3^2, 128\times k \\
1\times1^2, 128\times k 
\end{bmatrix} \times 3 $, 
\end{tabular} & 
{\begin{tabular}[c]{@{}l@{}} 
\textit{RN}: $16\times8^2$\\
\textit{BN}: $29\times8^2$
\end{tabular}}\\ \hline

conv4\_x & 
\begin{tabular}[c]{@{}l@{}} $ \begin{bmatrix}
1\times1^2, 512\\
3\times3^2, 256\\
1\times1^2, 256
\end{bmatrix} $, \\
$ \begin{bmatrix}
1\times1^2, 1024\\
3\times3^2, 256\\
1\times1^2, 256
\end{bmatrix} \times 5 $
\end{tabular} & 
\begin{tabular}[c]{@{}l@{}} $ \begin{bmatrix}
1\times1^2, 512\times k \\
\textcolor{BN1}{1}/\textcolor{BN9}{1}/\textcolor{BN17}{3}/\textcolor{BN33}{3}\times3^2, 256\times k \\
1\times1^2, 256\times k 
\end{bmatrix}$,  \\
$ \begin{bmatrix}
1\times1^2, 1024\times k \\
\textcolor{BN1}{1}/\textcolor{BN9}{1}/\textcolor{BN17}{1}/\textcolor{BN33}{1}\times3^2, 256\times k \\
1\times1^2, 256\times k 
\end{bmatrix} \times 5 $
\end{tabular} 
& 
{\begin{tabular}[c]{@{}l@{}} 
\textit{RN}: $8\times4^2$\\
\textit{BN}: $14\times4^2$
\end{tabular}}\\ \hline

conv5\_x &
\begin{tabular}[c]{@{}l@{}} $ \begin{bmatrix}
1\times1^2, 1024\\
3\times3^2, 512\\
1\times1^2, 512
\end{bmatrix} $, \\
$ \begin{bmatrix}
1\times1^2, 2048\\
3\times3^2, 512\\
1\times1^2, 512
\end{bmatrix} \times 2 $
\end{tabular} & 
\begin{tabular}[c]{@{}l@{}} $ \begin{bmatrix}
1\times1^2, 1024\times k \\
\textcolor{BN1}{1}/\textcolor{BN9}{1}/\textcolor{BN17}{1}/\textcolor{BN33}{3}\times3^2, 512\times k \\
1\times1^2, 512\times k 
\end{bmatrix} $, \\
$ \begin{bmatrix}
1\times1^2, 2048\times k \\
\textcolor{BN1}{1}/\textcolor{BN9}{1}/\textcolor{BN17}{1}/\textcolor{BN33}{1}\times3^2, 512\times k \\
1\times1^2, 512\times k 
\end{bmatrix} \times 2 $
\end{tabular}&
{\begin{tabular}[c]{@{}l@{}} 
\textit{RN}: $4\times2^2$\\
\textit{BN}: $6\times2^2$
\end{tabular}}\\ \hline

         & \multicolumn{2}{c}{Average pool, n\_classes-d fc, softmax} &  \\ %\thickhline
\end{tabular}
\caption{%\jvg{In principle I like the table; but I find it a bit difficult to interpret} 
Network architectures: 3D ResNet-50 (\textit{RN}) vs Video BagNet-\textcolor{BN1}{1},  \textcolor{BN9}{9}, \textcolor{BN17}{17} and \textcolor{BN33}{33} (\textit{BN}). 
In the first row, we report the number of parameters. The next rows correspond to the network layers, which contain convolutions and downsampling. For the convolutional layers, we report the kernel size $T\times S^2$, in the temporal ($T$) and spatial ($S^2$) dimensions, and the number of channels. The rightmost column of the table reports the output sizes at each layer, given an input clip of size $64\times64^2$. %This input size is large enough to show the effect of down-sampling. 
The convolutional blocks follow the structure of ResNet Bottleneck blocks \cite{he2016deep}. We widen the channels of Video BagNet with factor $k$, equal to \textcolor{BN1}{1.40}, \textcolor{BN9}{1.40}, \textcolor{BN17}{1.35} and \textcolor{BN33}{1.25},
%\klamer{But this widening is not in this table?} 
to keep the number of parameters comparable among the different models. 
In both architectures, each %convolutional 
layer is followed by Batch Norm \cite{ioffe2015batch} and a ReLU \cite{nair2010rectified}. % \klamer{we talk about long-term videos -- so 64 input frames are not really representative? Explain why 64 is taken}\om{todo: mention 64 is enough to show the effect of down-sampling. Add Video BagNet-1.}
}
\label{tab:architecture}
\end{table*}

\section{Method}
\label{sec:method}

We study how the size of the temporal RF effects model sensitivity to sub-action order. %\klamer{and length}. 
To this end, we compare long-term action recognition performance of 3D convolutional networks with variable temporal RF size. %\om{Say why we only focus on convolution and do not use Transformers}

\subsection{Video BagNet}

Inspired by the 2D BagNet for image classification~\cite{BagNet}, we design Video BagNet, a 3D convolutional network that 
%is capable to reason over short temporal extents \rb{Conventional large RF models are also "capable" to reason over short temporal extents, I guess? Video BagNet seems uniquely restricted to short extents?}. 
reasons over short temporal extents.
The key idea behind Video BagNet is to harness bag-of-feature representations for video classification. Specifically, the word vocabulary is composed of short video segments. Although this representation does not allow to model long-term temporal dependencies, it prevents learning strict temporal orders that can lead to the misclassification of a video if %random \klamer{unseen (they don't have to be random...} 
unseen permutations between sub-actions occur at test time.

Our Video BagNet is based on the 3D ResNet-50 %utilized \klamer{described / introduced} 
described in Hara \etal{} \cite{hara2018can}. We apply a set of modifications to %the latter network \rb{no references, just repeat "3D ResNet-50" (Jan's guidelines)}
3D ResNet-50 to restrict the size of its temporal receptive field, while leaving the computation in the spatial dimensions unchanged. In particular, we propose four variants of Video BagNet, with temporal RF sizes of 1, 9, 17, and 33 input frames. We choose these temporal extents following the design choice of Brendel \etal{} \cite{BagNet} in the image domain. Video BagNet is sensitive to order within its small temporal RF, allowing for fine-grained motion modeling. %\rb{This last sentence is tackling a concern the reader might have, it seems? Maybe describe the concern explicitly too, not just the solution?}

\begin{figure}[]
\centering
\includegraphics[width=0.9\linewidth]
{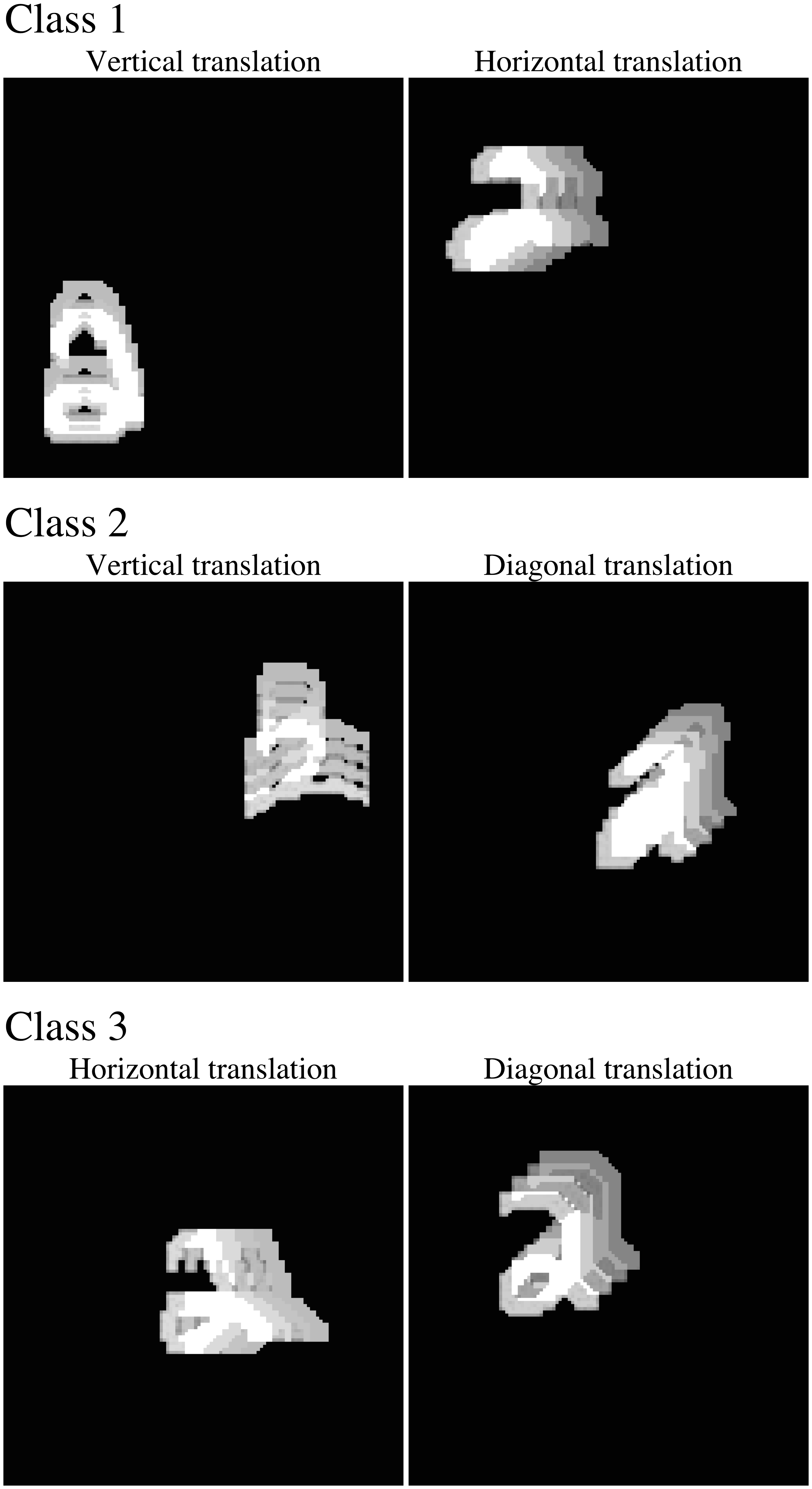}
\caption{%\jvg{Very nice figure} 
Example of videos of digit $2$ from the \textit{Directional Moving MNIST} dataset. The videos are composed of two sub-actions, i.e. vertical, horizontal or diagonal translation. Sub-action co-occurrences determine the video class. We explicitly superimposed multiple frames with shading to show the movement.
%\rb{Do you want to say explicitly that you've superimposed multiple frames with shading? I don't know if this is common in video literature...}
\vspace{-0.2cm}
}
\label{fig:directional_moving_MNIST}
\end{figure}

% The main differences \klamer{this makes me question what other modifications there are} 
% The differences between Video BagNet and 3D ResNet-50 can be summarized as follows.
The set of modifications that we apply to 3D ResNet-50 can be summarized as follows.

First, we restrict the size of some of the convolutional kernels in the temporal dimensions. This is done to adaptively control the expansion of the RF in the temporal dimension through the convolutional layers, without changing the depth of the network. We express the size of the convolutional kernels in the temporal ($T$) and spatial ($S^2$) dimensions as $T\times S^2$.
The $7\times7^2$ convolutional kernel in the first layer is replaced with a convolutional kernel of size $3\times7^2$ ($1\times7^2$ for Video BagNet-1). In the following layers, we modify a set of 3D ResNet-50 bottleneck blocks. Bottleneck blocks consist of three consecutive convolutional layers of size 
$\begin{bmatrix}
1\times1^2,\\
3\times3^2,\\
1\times1^2
\end{bmatrix}$. 
We replace them with 
$ \begin{bmatrix}
1\times1^2,\\
1\times3^2,\\
1\times1^2
\end{bmatrix}$.  

In addition, to prevent the temporal RF size from growing in the first layer, we alter the MaxPool operator that follows layer \textit{conv1} to perform pooling only in the spatial dimensions. 
To maintain a comparable amount of parameters between 3D ResNet-50 and the different Video BagNet models, we widen the number of channels. Finally, to keep the input size equal to the video length, we remove the padding.
% An overview of the architecture design of Video BagNet %\rb{Video or video? I changed it to capital in a couple of places already...} 
% and the differences from 3D ResNet-50 is provided in Table \ref{tab:architecture}. 
%\om{more info about channels, etc in the table}
% Architecture, modifications, analysis of RF size (maybe follow method in BagNet)
% Mathematical expression?
%
Table \ref{tab:architecture} provides an overview of the architecture design of Video BagNet vs. 3D ResNet-50.

\section{Experiments}
\label{sec:results}
\subsection{Datasets}
%\jvg{in this case it seems better to not have first a 'datasets' part, but to put the datasets together with the experimental question you aim to answer. }
We study the effect of the temporal RF size on two long-term datasets, namely the \textit{Directional Moving MNIST}, that we propose, and MultiTHUMOS \cite{MultiTHUMOSDataset}. %\xin{is it long term?}\om{yes, because the videos are composed of multiple sub-actions}, 
%\klamer{that we propose, and MultiTHUMOS \cite{MultiTHUMOSDataset}}. 
These datasets contain multiple sub-actions and can last up to several minutes. %In this case \klamer{which of the two, or both?}
%\rb{I see this paragraph has been edited already, but this sentence "Long-term..." doesn't naturally connect to the previous sentence. Maybe something like "These datasets consist of long-term videos that can last up to several minutes and have multiple sub-actions."}
For these datasets, the classification task consists of recognizing the sub-actions that compose the videos. 

\textbf{Directional Moving MNIST} %\xin{Can we give some example images of this dataset?}
is a dataset  %\rb{is the dataset long-term? "Dataset of long-term videos" or something like that? What makes the dataset "long-term"?}
composed of videos of one single moving digit, %\klamner{image ?}, 
randomly sampled from the original MNIST dataset \cite{deng2012mnist}. It contains 3 classes and 1000 videos per class. 
%\rb{Up to this point, it is not very clear to me that each sample has multiple subsequent translations. So the next sentence reads a bit confusion, as it mentions multiple translations.}
% Three possible sub-actions occur: vertical translation, horizontal translation and diagonal translation.
In this dataset, the digit translations correspond to sub-actions and the co-occurrence of two sub-actions determines the video class. More specifically, vertical and horizontal translation form class 1, vertical and diagonal translation form class 2 and horizontal and diagonal translation form class 3. 
%\klamer{what is varied between samples within a class? Isn't that motion amplitude / speed / frequency?}
%
Within each class, digit appearance and starting position have been randomized.
%\rb{simpler: "digit appearance and starting position has been randomized."}. 
In addition, the translations occur at two possible speeds.
All sub-actions have equal duration and there are no pauses between consecutive sub-actions. 

One fixed sub-action order appears in the training set. At test time we use two sets: in the \textit{test set without permutations}, the sub-action order is the same as training time; while in the \textit{test set with permutations} the sub-action order is permuted with 50\% probability.
An example of the \textit{Directional Moving MNIST} dataset is provided in \fig{directional_moving_MNIST}. % \klamer{missing figure numnber}

\begin{figure*}[]
	\centering
	\begin{tabular}{c@{}c}
	\includegraphics[width= 0.45\linewidth]
	{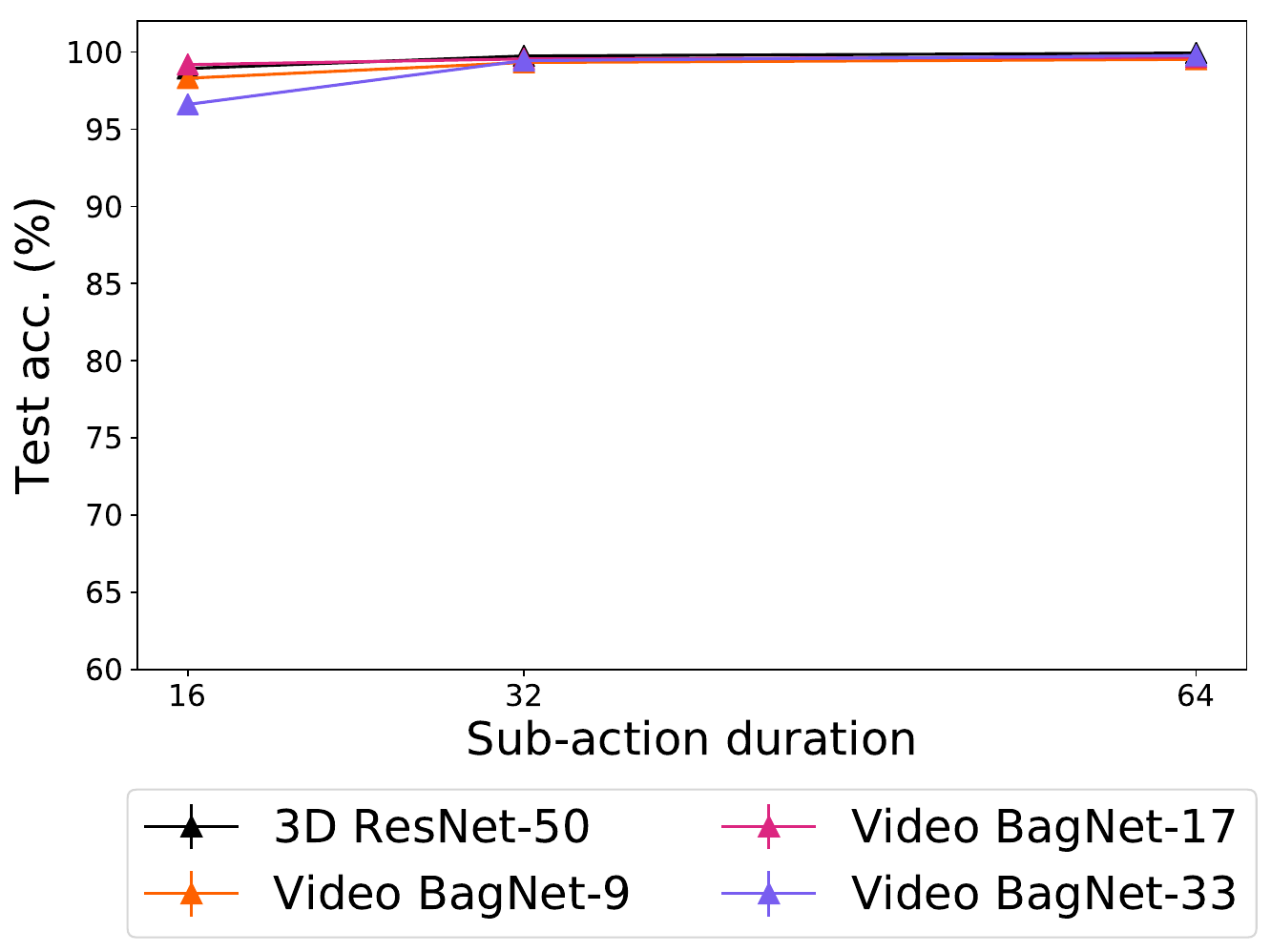} & 
	\includegraphics[width= 0.45\linewidth]
	{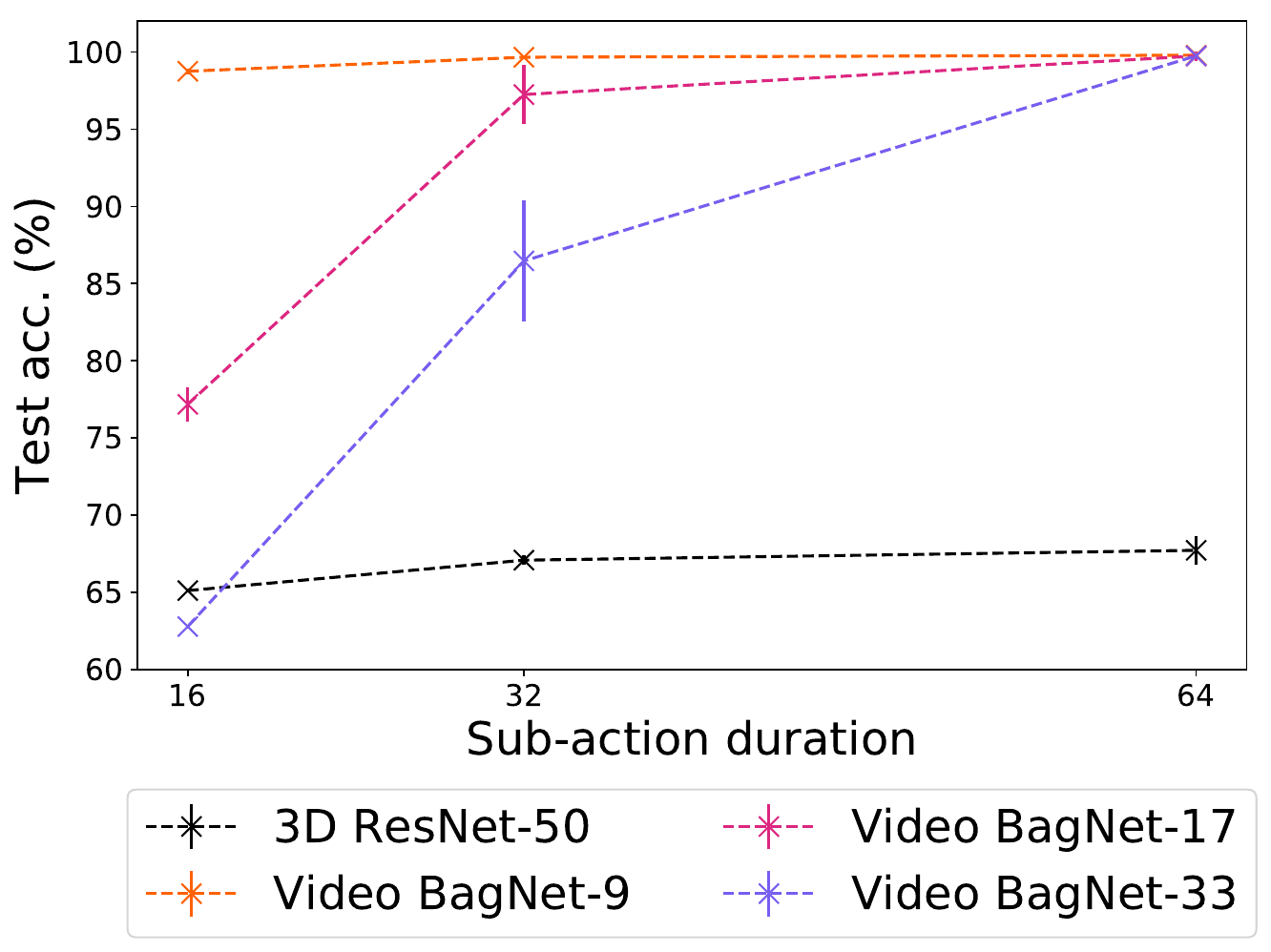} \\
	(a) Test set with no permutations & (b) Test set with permutations
	\end{tabular}
	\setlength{\belowcaptionskip}{-8pt}
	\caption{%\jvg{Include info on which dataset this is} 
	%\jvg{marker size bit larger} 
	Sensitivity to sub-action order on the \textit{Directional Moving MNIST} dataset. Models with different temporal RF are tested on two test sets with the same order (a) and different order (b) w.r.t. training time. The models with small temporal RF compared to the sub-action duration, namely Video BagNet 9, 17 and 33, perform well on the two sets. Differently, 3D ResNet, with temporal RF larger than 100 frames, overfits the temporal order at training time and fails to classify the test set with permutations. %\klamer{is the sub-action duration the same during trainng and testing?}\om{add video bagnet-1}
	}
	\label{fig:toy_exp}
    % \vspace*{-3mm}
\end{figure*}

\textbf{MultiTHUMOS} \cite{MultiTHUMOSDataset} is a multi-label video dataset for long-term action recognition. It is a collection of 400 complex, unconstrained, sports videos that have been densely annotated with sub-action time steps. The dataset contains a total of 65 possible sub-actions and each video contains, on average, $84.03 \pm 113.56$ sub-actions. The small size of the dataset prevents from training classification models using all the possible sub-action combinations and orders that usually occur in sports videos. For example, the dataset contains 20 basketball videos of which 15 videos contain the sub-actions \textit{BasketballDribble}, \textit{Run}, \textit{BasketballPass}. Only 4 videos contain the order \textit{BasketballDribble - Run - BasketballPass}.
%\rb{I feel like this is a pretty important point, but the way you make it is kind of "handwavy". I'm not saying it isn't true, but I feel like I just "have to trust you" on it. Is there some way to quantify this? Or maybe it could be made more obvious from an example like "for tennis, we have 10 videos, which only show five different kinds of points."}
% \om{todo: add example for a few different video annotations}

\subsection{The size of the temporal RF affects model sensitivity to sub-action order}
% \subsection{Large temporal RF are sensitive to sub-action order \xin{Temporal RF size effects the sensitivity of model to sub-action orders}}
We design a simple controlled experiment to investigate whether spatio-temporal models encode the sub-action order through their temporal RF. For this, we deploy the \textit{Directional Moving MNIST} dataset.
We vary the size of sub-actions to relate it with different temporal RF sizes. Specifically, we use sub-action duration of 16, 32 or 64 frames and temporal RF size equal to 217 frames for 3D ResNet-50 and 9, 17 and 33 frames for our Video BagNet. %We use a widening factor to the Video BagNet channels to maintain the number of parameters comparable to 3D ResNet-50. \klamer{do not repeat that here, that is section 3 stuff}

The results of this experiment are summarized in \fig{toy_exp}. Irrespectively of the temporal RF size and the sub-action duration, all the models perform well when the order of sub-actions of the training and test sets match, that is in the \textit{test set without permutations}. However, on the \textit{test set with permutations}, the models with large temporal RF size compared to the sub-action duration, e.g. 3D ResNet-50, and, in some instances, Video BagNet-17 and Video BagNet-33, perform poorly. In particular, 3D ResNet-50 always achieves an accuracy of $\sim66\%$, which is equivalent to classifying correctly the videos with no permutations ($\sim50\%$ of the \textit{test set with permutations}) and randomly the videos with sub-action permutations. Our Video BagNet-9, which has the shortest temporal RF among the analyzed models, performs above $98.5\%$ on all the different test videos. 

\begin{figure}[]
\centering
\includegraphics[width= 0.9\linewidth]
{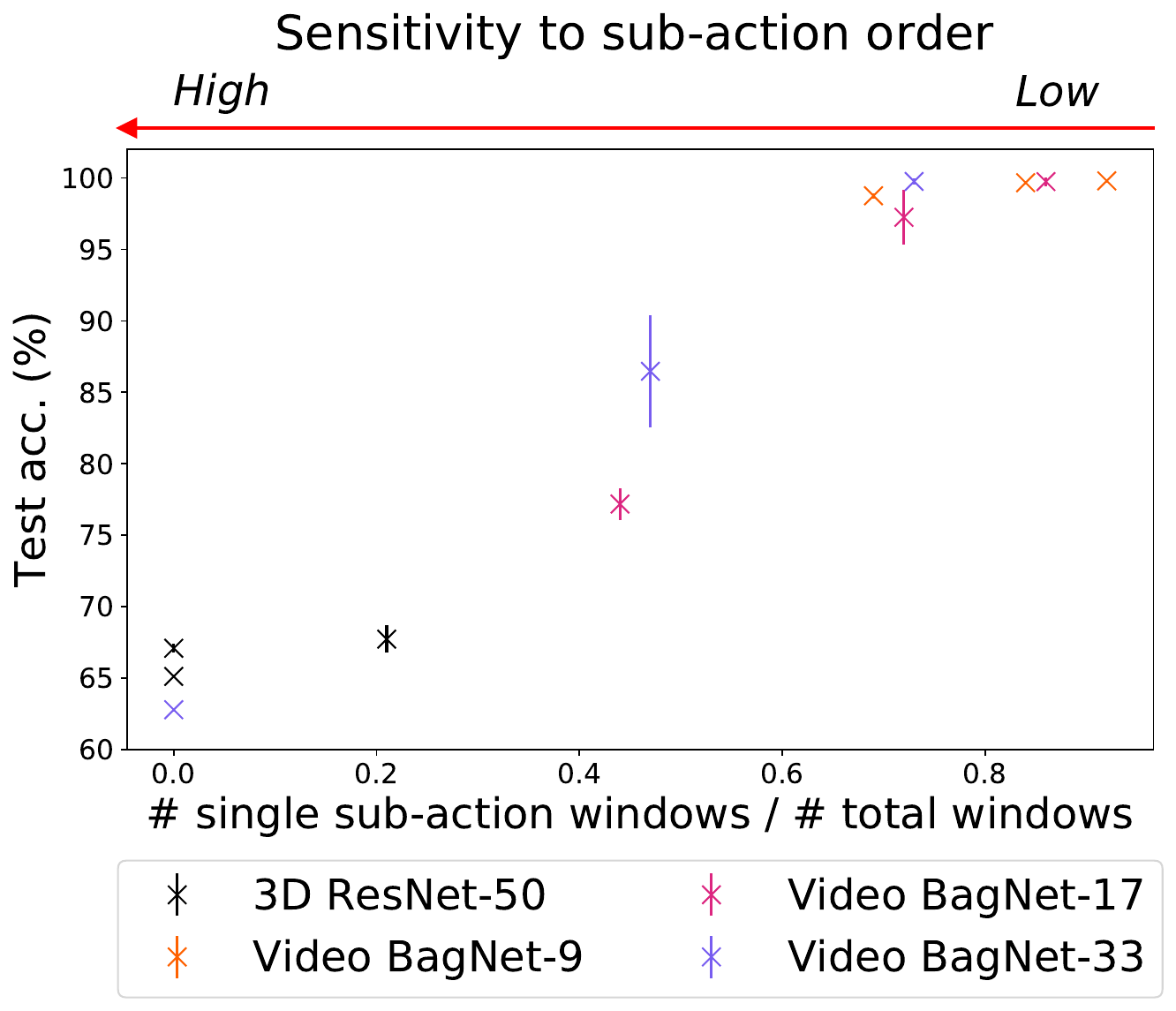}
\caption{%\jvg{increase size of marker} 
Accuracy on the \textit{Directional Moving MNIST} test set with permutations in terms of models sensitivity to sub-action order.
Sensitivity to sub-action order depends on the sub-action duration and temporal RF size, as shown in \fig{tRF_comparison}. It can be expressed by counting the amount of convolutional shifting windows that contain single sub-actions (\textit{\# single sub-action windows}) over the total convolutional windows (\textit{\# total windows}). Models with large ratio \textit{\# single sub-action windows / \# total windows}, like Video BagNet-9, are less sensitive to order and achieve good performance.
Models with very large temporal RF sizes, like 3D ResNet-50, always see sub-action co-occurrences rather than single sub-actions. Therefore, their ratio \textit{\# single sub-action windows / \# total windows} is low and their order sensitivity is high, thus performing poorly on the test set with permutations. 
%\om{refer to figure1}
}
%\xin{Is it possible to put 'high','low' text on the arrow of sensitivity?}

\label{fig:single_window}
\end{figure}

%\rb{Separate subsection for these results?} 
These results show that sensitivity to sub-action order depends on the sub-action duration and temporal RF size. We quantify the sensitivity to order by relating the sub-action size to the temporal RF size. 
For this, we analyze the convolutional shifting windows in the last convolutional layer of the 3D ResNet-50 and Video BagNet models, represented in \fig{tRF_comparison}.
In particular, %we count the amount of convolutional
we measure the sensitivity by a ratio of the amount of shifting windows that contain single sub-actions (\textit{\# single sub-action windows}) over the total amount of convolutional windows (\textit{\# total windows}). When the ratio is high,
%the convolutional windows with single sub-actions are significantly more frequent than that containing sub-actions co-occurrences, 
the sensitivity to the sub-action order is low.
As shown in \fig{tRF_comparison},  models with very large temporal RF size, like 3D ResNet-50, always see sub-action co-occurrences rather than single sub-actions. Therefore, in \fig{single_window}, their ratio \textit{\# single sub-action windows / \# total windows} 
% \jvg{needs a bit more explanation} 
is always low, which leads to low performance on the test sets with permutations.
On the other hand, models with small temporal RF size, e.g. Video BagNet-9, have a large ratio of \textit{\# single sub-action windows / \# total windows} and low sensitivity to the sub-action order, achieving good performance on the test set with permutations. 
%\klamer{basically we show here that we should match RF size to subaction size, rather than something specific on order...}

\subsection{Small vs. large temporal RF for long-term video action recognition} 

\begin{table}[!ht]
\centering
\begin{tabular}{lll}
\hline
Model           & Temporal RF       & mAP   \\ \hline

Single-frame CNN \cite{simonyan2014very} & 1          & 25.4 \\
% Two-stream CNN \cite{} & ?          & 27.6 \\
MultiLSTM \cite{MultiTHUMOSDataset} & 15                & 29.7 \\
\midrule
3D ResNet-50 \cite{hara2018can}   & \textgreater{}100 & 22.45 \\

Video BagNet-33 & 33                & 26.37 \\
Video BagNet-17 & 17                & 28.97 \\
Video BagNet-9  & 9                 & 30.21 \\
Video BagNet-1  & 1                 & 12.60 \\
\hline
\\
\end{tabular}
\caption{Classification accuracies of models with small and large temporal RF on the MultiTHUMOS dataset. We compared our evaluated models (bottom rows) to the baselines proposed in \cite{MultiTHUMOSDataset} (top rows). Despite being trained from scratch, our Video BagNet models with temporal RF 9, 17 and 33 perform comparably to the ImageNet \cite{russakovsky2015imagenet} pre-trained baselines.
Models with smaller temporal RF, e.g. Video BagNet-9, recognize sub-action occurrences and ignore temporal order, achieving the best performance. Video BagNet-1 cannot model motion by seeing just single frames, which has the lowest mean average precision.}
\label{tab:MultiTHUMOS}
\vspace*{-3mm}
\end{table}

In our controlled experiment, we show that models with large temporal RF encode 
%\rb{"model" or "overfit to"? Would be more consistent with earlier language} 
the sub-action order at training time. We argue that this causes misclassification when the distributions of sub-actions order are different in the training and test sets. This is the case for the commonly used MultiTHUMOS dataset, which only consists of 400 videos with high variability in sub-actions composition and order. 
%\rb{Same point as I made earlier, I guess, but: do you have any proof for that high variability?}

We evaluate the effect of the temporal RF size on %this dataset \rb{Jan's guidelines: don't refer, repeat. "on MultuTHUMOS"}
MultiTHUMOS. Again, we deploy
%\rb{"Again, we deploy"} 
3D ResNet-50 and Video BagNet with temporal RF 1, 9, 17 and 33. We train the models from scratch, without using either pre-training or data augmentation. 
%\rb{"using neither pre-training nor data augmentation" https://www.quickanddirtytips.com/education/grammar/when-use-nor (I also had to look this up)}. 
We train with 512 input frames, with batch size 4. 
% \om{how are the frames sampled?}
We do this to limit the computational effort of our experiments. 
Since we train the models from scratch and without data augmentation, our results are not comparable to current state-of-the-art \cite{zhou2021graph}. Nevertheless, employing this fixed experimental setup for all the analyzed models allows us to fairly compare different temporal RF sizes. 
%\rb{Why? I presume you want to claim that what you test is not affected by the limited setting, but in the current writing the reader has to infer this themselves. For example, I am not enough of a video expert to know this, so I'm left having to trust you on it.}

The results in \tab{MultiTHUMOS} show that models with small temporal RF size outperform models with large temporal RF size on this dataset. The highest accuracy is obtained with Video BagNet-9. %, which has the smallest temporal RF size. 
These results suggest that encoding long-term information, including sub-action order, %is not necessary for \klamer{is hurting} 
is hurting the classification of MultiTHUMOS. 
This long-term information could correspond to the precise order of sub-actions or to the varying durations of different sub-actions.
This is sensible: the multi-label classification problem of MultiTHUMOS consists in recognizing all the single sub-actions occurring in a video. Sub-action classification can be achieved by looking at short temporal extents that contain the sub-action. Because of the high variation in the temporal composition of sports videos, overemphasizing long-term information is not necessary or even decreases the sub-action recognition accuracy.
On the other hand, for Video BagNet-1 it shows that if the model encodes neither long-term nor short-term information, the accuracy decreases.
The results indicate that the short-term information captured by small temporal RF seems essential for good classification performance.
%\klamer{To make that statement you should compare to a setup without any short-term information, such as a single-frame classifier...}\om{todo: add Video BagNet-1}. 
%
 % \om{Speculate that effect might be due to multiple factors, e.g. duration, or order.}
%
% Overemphasizing long-term information is not necessary or even decrease the action recognition accuracy. \om{Need to rephrase this in a better way}
%
%
%The conclusions from this experiment are in contrast to the current direction of long-term activity recognition research, which builds towards large or global temporal RF \om{todo: add citations}. \xin{I think this part is left to BMVC submission?} \rb{I would personally put these kind of more "discussion" comments in the Conclusion/Discussion. It fits well in a section that discusses future work, for example.}

%Due to our limited computational resources, the lack of pre-training makes it impossible to compare our results to the current state-of-the-art on MultiTHUMOS. Hence, we 
%Comparing our work to the baseline models proposed in \cite{MultiTHUMOSDataset}, 
We find that our results are comparable % show similar performance 
to the baseline models proposed in \cite{MultiTHUMOSDataset}, as illustrated in Table \ref{tab:MultiTHUMOS}. 
It is worth noting that the single-frame CNN \cite{simonyan2014very}, which cannot model temporal information by design, has the advantage of being pre-trained on ImageNet \cite{russakovsky2015imagenet}, thus explaining the superior performance compared to Video BagNet-1. 
Similarly, the MultiLSTM model \cite{simonyan2014very} uses pre-trained image features. Despite the lack of pre-training, Video BagNet-9 and 17 achieve 28.97\% and 30.21\% mAP, which is similar to mAP of 29.7\% mAP obtained by Video MultiLSTM.

% MultiLSTM finds temporal relationships over a window of 15 frame features, 
% achieving 29.7\% mAP, which is similar to mAP of 28.97\% and 30.21\%, obtained by Video BagNet-9 and 17. 

\section{Conclusions}
\label{sec:discussion}

In this paper, we investigate whether spatio-temporal models for long-term action recognition encode sub-action order through their temporal RF. Our experiments reveal that when the temporal RF size is larger than the sub-action duration, the models are sensitive to the sub-action order. We provide a measure for the sensitivity to the sub-action order by a ratio of the number of convolutional windows that contain single sub-actions over the total number of convolutional windows. A higher ratio makes the models less sensitive to the sub-action order. 
%\klamer{bad sentence -- rephrase}.

% We show in a fully-controlled experiment that 
Sensitivity to sub-action order causes misclassification when the order 
%\klamer{and length} 
of sub-actions are different during training and test time. 
%
%
% This occurs in both synthetic and real world long-term action recognition datasets. 
This might occur in long-term action recognition, since it is difficult to collect training samples containing all the % possible 
sub-action permutations that exist in natural videos.
We show that small temporal RFs are robust to permutations of sub-actions, which is beneficial when limited sub-action orders are available at training time.
% In addition, small temporal RFs outperform large temporal RFs on the MultiTHUMOS dataset. We hypothesize that this is due to short temporal RFs being less sensitive to specific long-term information, like sub-action order or duration.
%
% Using short temporal RFs is beneficial when limited training data is available, sub-action orders are available at training time, as in the MultiTHUMOS dataset.
%
%
% We recommend to adopt short temporal receptive fields when it is not possible to train with %all possible \klamer{a full representative set of} 
% a full representative set of sub-action orders. 
%An alternative strategy can be performing data augmentation and generate new training orders. We will investigate this strategy in future work. \xin{Maybe this paragraph can be removed?} \klamer{keep the first line, perhaps delete the alternative}
%
Our study is conducted on 3D convolutional networks. Nevertheless, the conclusions could be generalizable %\rb{"could"? Otherwise it sounds a bit too speculative, for my personal taste.} 
to other spatio-temporal models that use the RF to encode temporal dependencies. 
%e.g. Transformers. We will include experiments with Transformers in future work. \rb{Jan's guidelines: don't list "experiments we didn't do" as future work.}

% \small
\smallskip\noindent\textbf{Acknowledgements.} 
This work is part of the research program Efficient Deep Learning (EDL), which is (partly) financed by the Dutch Research Council (NWO).

{\small
\bibliographystyle{ieee_fullname}
\bibliography{egbib}
}

\end{document}